\documentclass[letterpaper, 10 pt, conference]{ieeeconf}  
\usepackage[pdftex]{graphicx}
\usepackage{subcaption}
\usepackage{hyperref}

\IEEEoverridecommandlockouts                              

\title{\LARGE \bf
Rotational Slippage Prediction from Segmentation of Tactile Images
}

\author{Julio Castaño-Amorós$^{1}$ and Pablo Gil$^{2}$
\thanks{*This research was funded by the Valencian Regional Government through the PROMETEO/2021/075 project and by the University of Alicante through the grant UAFPU21-26
}
\thanks{$^{1}$Julio Castaño-Amorós is with University Institute for Engineering Research, Miguel Hernández University, Elche and with AUROVA Lab, Computer Science Research Institute, University of Alicante, Alicante, Spain
        {\tt\small julio.ca@ua.es}}%
\thanks{$^{2}$Pablo Gil is with AUROVA Lab, Computer Science Research Institute and with Department of Physics, Systems Engineering, and Signal Theory, University of Alicante, Alicante, Spain
        {\tt\small pablo.gil@ua.es}}%
}

\begin{document}

\maketitle
\thispagestyle{empty}
\pagestyle{empty}

\begin{abstract}
Adding tactile sensors to a robotic system is becoming a common practice to achieve more complex manipulation skills than those robotics systems that only use external cameras to manipulate objects. The key of tactile sensors is that they provide extra information about the physical properties of the grasping. 
In this paper, we implemented a system to predict and quantify the rotational slippage of objects in hand using the vision-based tactile sensor known as Digit. Our system comprises a neural network that obtains the segmented contact region (object-sensor), to later calculate the slippage rotation angle from this region using a thinning algorithm.
Besides, we created our own tactile segmentation dataset, which is the first one in the literature as far as we are concerned, to train and evaluate our neural network, obtaining results of 95\% and 91\% in Dice and IoU metrics. In real-scenario experiments, our system is able to predict rotational slippage with a maximum mean rotational error of 3 degrees with previously unseen objects. Thus, our system can be used to prevent an object from falling due to its slippage.
\end{abstract}

\section{Introduction and Related Work}
Traditionally, the methods to carry out robotic manipulation tasks used 2D or 3D vision sensors \cite{du}, which only take into account the geometric properties of the objects to perform the grasping. In contrast, with tactile sensors, it is possible to measure and react to physical properties (mass distribution, center of gravity or friction) in order to achieve a stable grasping \cite{luo}.
 
In the last twenty years, several tactile sensors have been designed using different hardware technologies \cite{chi}, although
the last trend of tactile sensors lies in optical tactile sensors \cite{hardwaretactil}.
In this manuscript, we present an algorithm to estimate the rotation angle of an object which is being manipulated when slippage occurs. This method is based on segmentation neural networks to obtain the contact region (object-sensor) and traditional computer vision techniques to calculate the rotation angle and is applied to the vision-based tactile sensor Digit \cite{digit} which does not contain visual markers to keep low its cost.

Estimating the contact region between the robot’s fingertips and the grasped object has been attempted to be solved in different ways.
For example, by subtracting contact and no-contact tactile images \cite{pytouch}, detecting and grouping visual markers \cite{ito}, throughout 3D reconstruction and photometric algorithms \cite{gelsight}, or using neural networks \cite{bauza}, \cite{lepora}.
In contrast, although our work is inspired by these previous articles, the main differences lie in the fact that we use the Digit sensors, without markers \cite{ito}, which do not produce depth information \cite{gelsight}, and state-of-the-art segmentation neural networks, which are more robust than subtracting operations \cite{pytouch} and vanilla CNN \cite{bauza}, and its training is more stable compared with GAN’s training \cite{lepora}.

Slippage is a common physical event that occurs during object manipulation, that has been tried to solve for several years employing different approaches. For example, detecting binary slippage events with traditional image preprocessing techniques \cite{riai}, combining convolutional and recurrent neural networks to classify slip in clockwise and counterclockwise rotation \cite{brayan} or estimating the slip rotation angle using vision-based tactile sensors with markers \cite{kolamuri} or force/torque sensors \cite{toskov}. In this paper, we have inspired our work in these methods that characterize and quantify the rotational slip. 

\section{Method}

We propose a two-stage method for touch region segmentation and rotational slippage prediction.
The first stage of our method is based on a segmentation neural network applied to vision-based tactile sensing, which we called Tactile Segmentation Neural Network (TSNN). In this work, our goal is only to segment the contact region, then we decided to use DeepLabV3+ \cite{deeplabv3+} architecture for experimentation. DeepLabV3+ is well-known for using an encoder-decoder architecture to perform image segmentation, and for introducing a new layer in its architecture, which is a combination of atrous or dilated and depth-wise separable convolutions. This combination leads to a reduction of computational complexity while maintaining similar or even better performance than previous versions. As the encoder, the authors used a modified version of the architecture Xception, called Aligned Xception, which replaces all the max pooling layers by the depth-wise separable convolutions to perform the feature extraction procedure. The decoder, in contrast, is a simpler part of the architecture, which only comprises convolution, concatenation, and upsampling layers to transform the intermediate features into the output.

\begin{figure}[htbp]
     \centering
         \centering
         \includegraphics[width=0.45\textwidth, height=12cm]{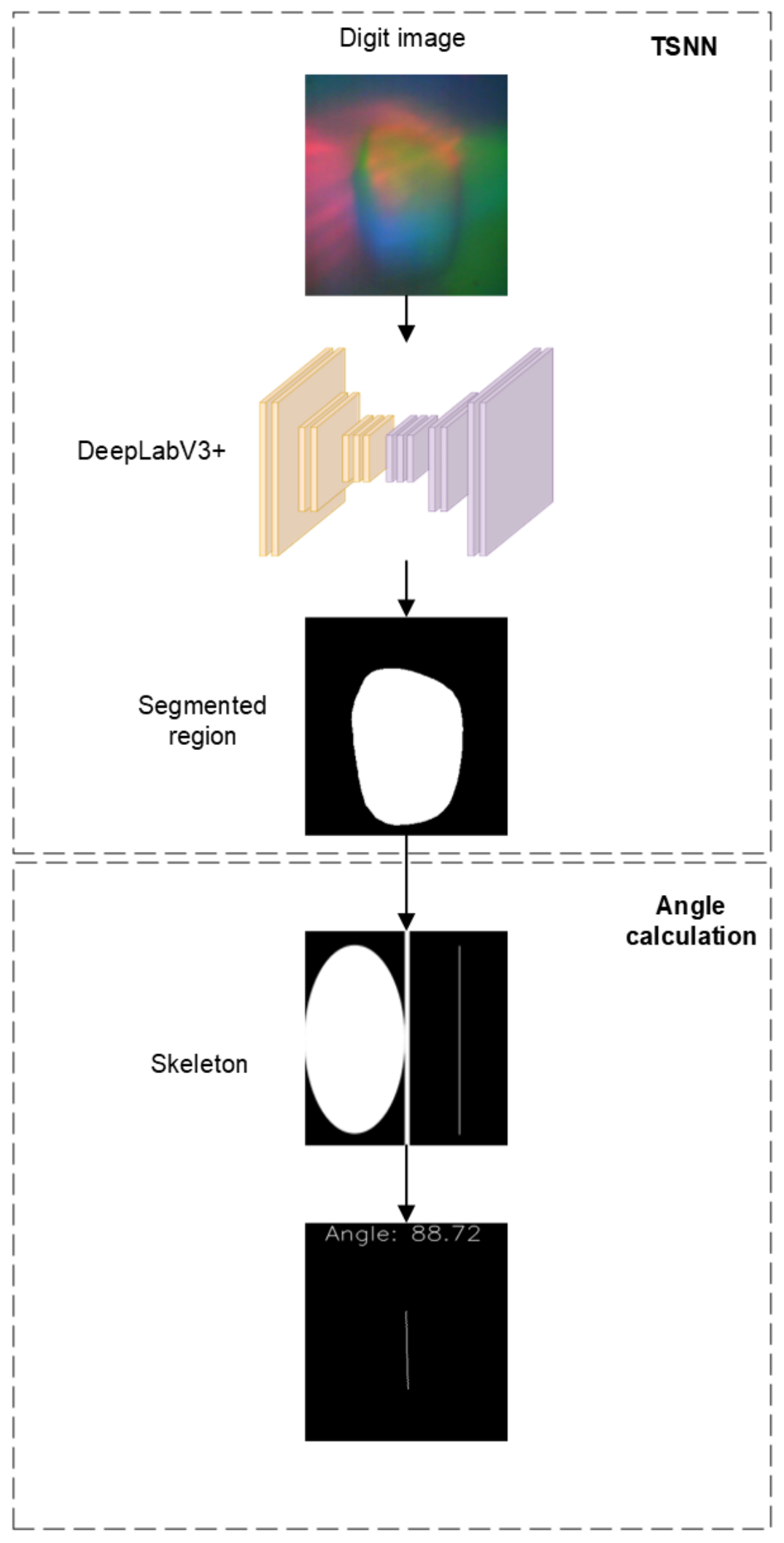}
     \caption{Diagram of our system combining both stages} 
     \label{fig:method}
\end{figure}

The second stage of our method estimates the angle of rotation of the segmented region 
of contact using a traditional computer vision thinning algorithm (Skeleton method) \cite{skeleton} that blackens points in the binary contact region using an 8-square neighborhood and different connectivity conditions. Other approaches, based on different neural networks such as Unet++ \cite{unet++} and PSPNet \cite{pspnet} or different algorithms to estimate the angle such as PCA or ellipse fitting, were tested. The complete system is shown in Fig \ref{fig:method}.


\section{Experimentation and results}
\label{section:experimentation}

We have generated our own dataset as we have not found any dataset related to tactile segmentation in order to be used as the base of our experimentation. Our tactile segmentation dataset comprises 3675 tactile images with their respective labelled contact regions. We have used 16 objects from YCB dataset to record it, from which we have captured between 200 and 250 tactile images per object. The objects contain different textures, rigidity, weight, geometries, etc. 

To train the TSNN we use the Dice and IoU metrics, an NVIDIA A100 Tensor Core GPU with 40 GB of RAM, and
the following optimal hyperparameters: a batch size of 32, a learning rate of 1e-4, the Adam optimizer, the Focal loss, and 30 training epochs. 

Table \ref{tab:tsnn_comparison} shows the results obtained by DeepLabV3+ TSNN in the testing experiment. DeepLabV3+ is able to segment tactile images with high accuracy and in real-time execution. Besides, this TSNN is 3 ms faster than other segmentation neural networks (Unet++ and PSPNet) while maintaining the same performance, thus, achieving a better trade-off between segmentation accuracy and prediction time. 

\begin{table}[htpb]
\centering
\caption{DeepLabV3+ TSNN results in terms of Dice, IoU and inference time metrics, and using the backbone ResNet18}
\label{tab:tsnn_comparison}

\begin{tabular}{|c|c|c|c|}
\hline
                    & \textbf{Dice}  & \textbf{IoU}   & \textbf{Time(s)} \\ \hline

\textbf{DeepLabV3+} & 0.956 $\pm$ 0.013 & 0.914 $\pm$ 0.023 & 0.006 $\pm$ 0.002   \\ \hline
\textbf{PSPNet} & 0.951 $\pm$ 0.014 & 0.907 $\pm$ 0.025 & 0.006 $\pm$ 0.002   \\ \hline
\end{tabular}
\end{table}



\begin{figure}[htbp]
     \centering
     \begin{subfigure}[b]{0.22\textwidth}
         \centering
         \includegraphics[height=3.8cm, width=4cm]{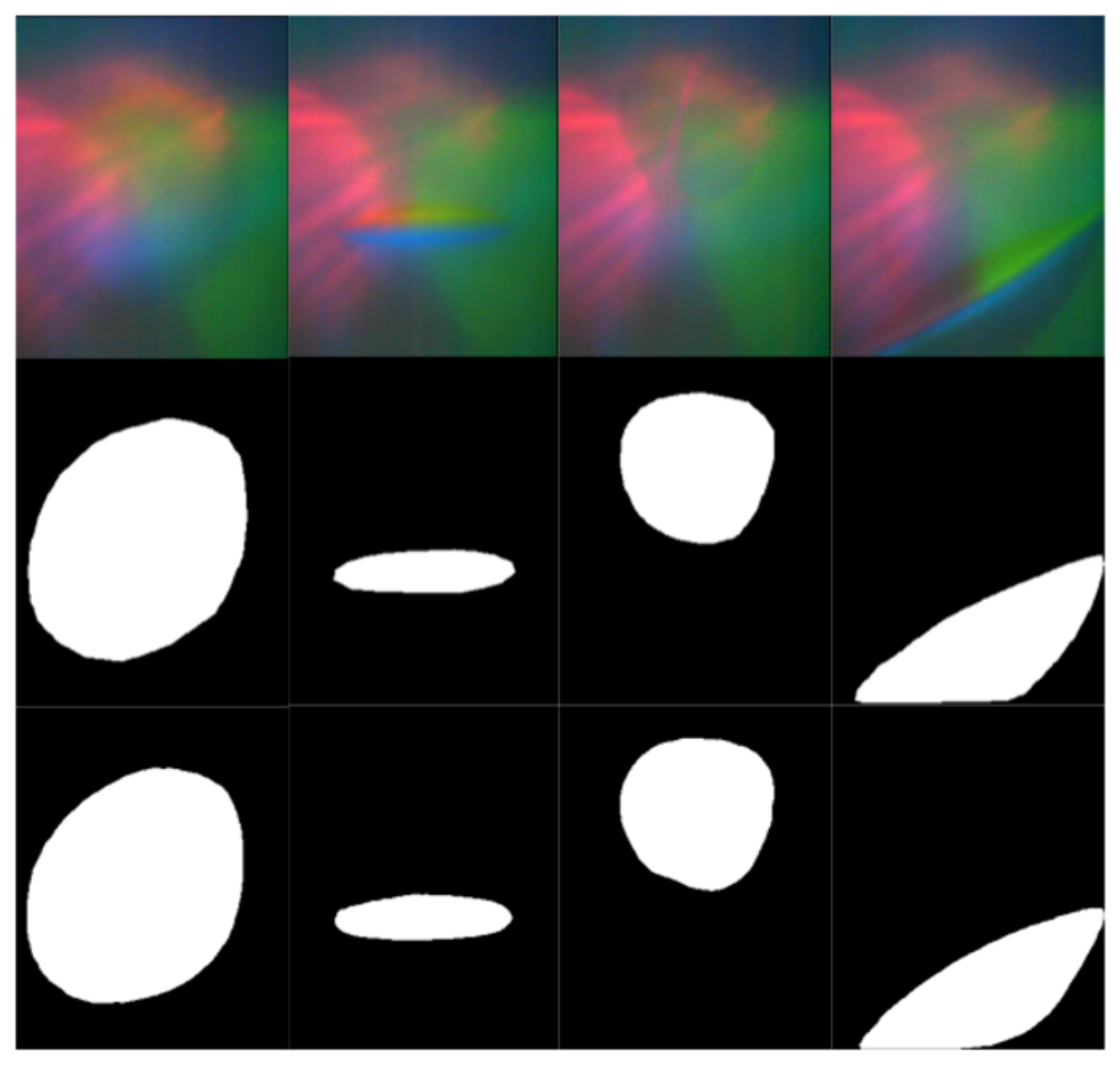}
         \caption{}
         \label{fig:examples_tactile_segmentation}
     \end{subfigure}
     \begin{subfigure}[b]{0.22\textwidth}
         \centering
         \includegraphics[height=3.8cm, width=4.3cm]{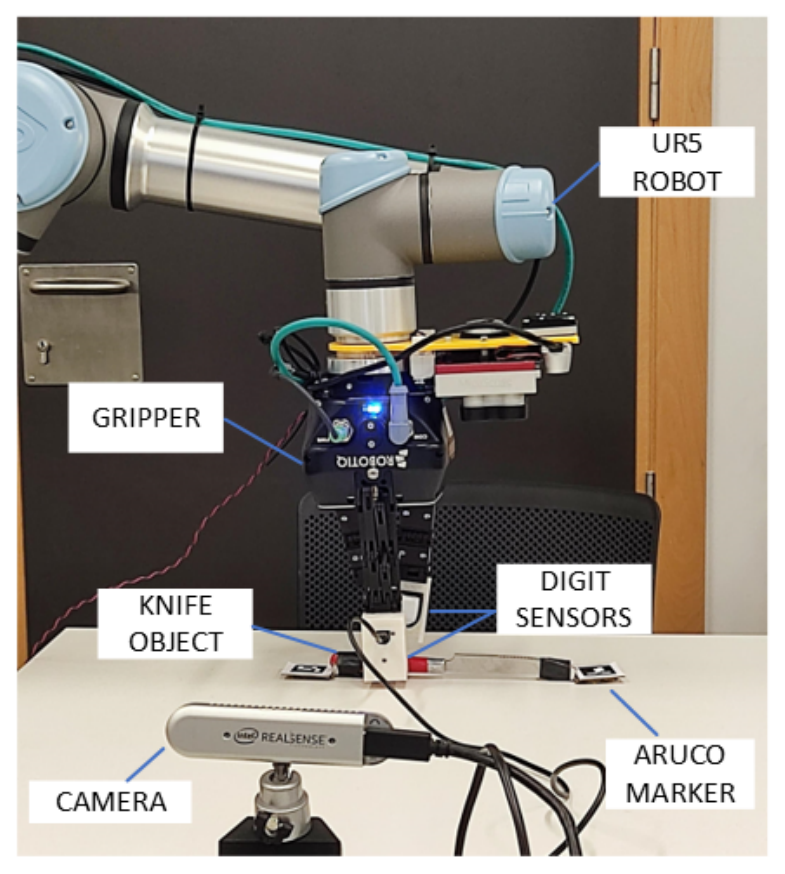}
         \caption{}
         \label{fig:setup}
     \end{subfigure}
     \caption{a) Examples of rotation angle calculation for slipping during lift task: DIGIT image (first row), ground truth (second row),  prediction (third row), b) Robotic manipulation setup
     with different objects} 
     
\end{figure}

Figure \ref{fig:examples_tactile_segmentation} shows different examples of contact region segmentation carried out by DeepLabV3+ TSNN, and Fig. \ref{fig:setup} shows our robotic manipulation setup with a UR5 robot, two DIGIT sensors, a 
ROBOTIQ gripper, the object to grasp with the aruco markers attached and an Intel RealSense camera to calculate the ground truth angle.

\begin{figure}[htbp]
     \centering
         \centering
         \includegraphics[width=0.45\textwidth,height=9cm]{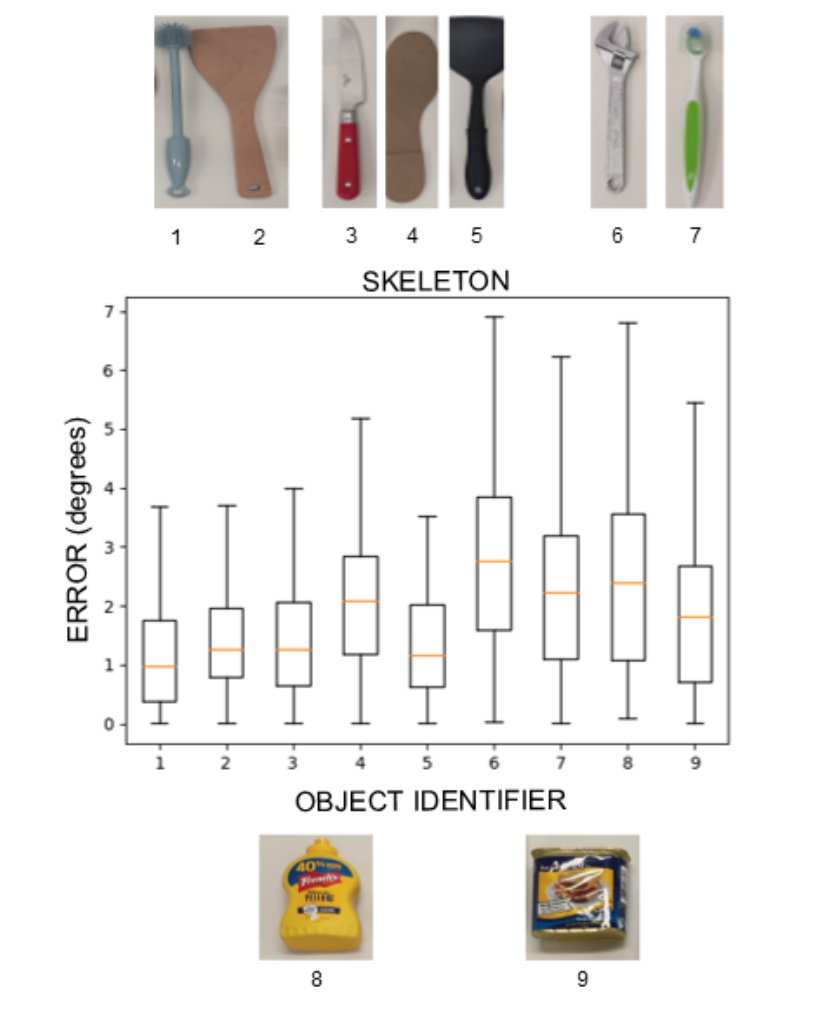}
     \caption{Rotation errors for each object using the Skeleton method} 
     \label{fig:skeleton_best_method}
\end{figure}

The task consists of grasping and lift an object while the tactile segmentation and rotational slippage angle are estimated.
The predicted angle is calculated as the difference between the current and the initial angle obtained in the Skeleton method described earlier, while the ground truth angle is calculated using two aruco markers as visual references. 
Our system was evaluated with seven unseen objects (1 to 7 in Fig. \ref{fig:skeleton_best_method}) and two seen objects from our tactile segmentation dataset (8 and 9 in Fig. \ref{fig:skeleton_best_method}).

\begin{figure}[htbp]
     \centering

     \begin{subfigure}[b]{0.22\textwidth}
         \centering
         \includegraphics[height=3cm]{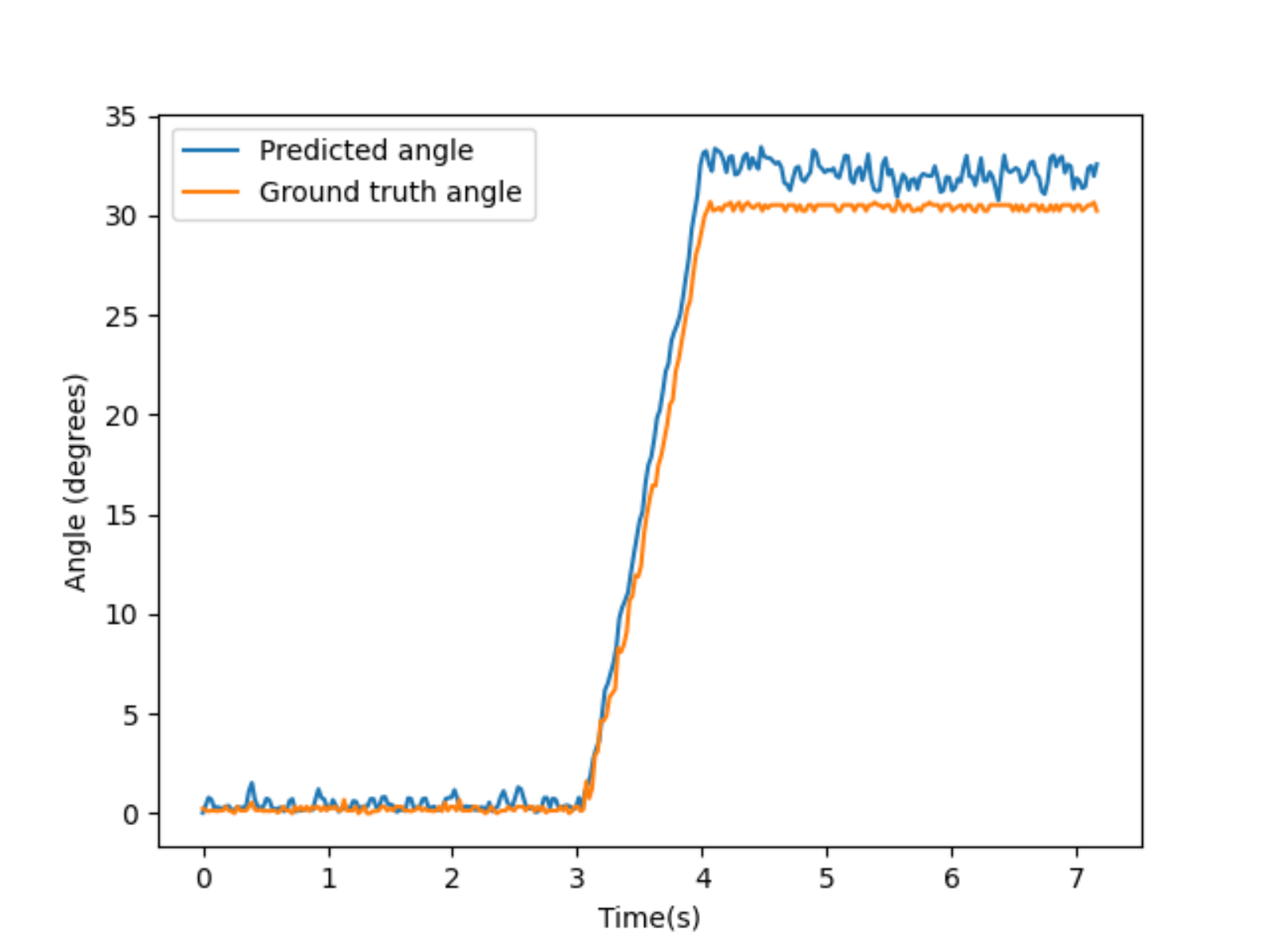}
         \caption{Object 3}
     \end{subfigure}
     \begin{subfigure}[b]{0.22\textwidth}
         \centering
         \includegraphics[height=3cm]{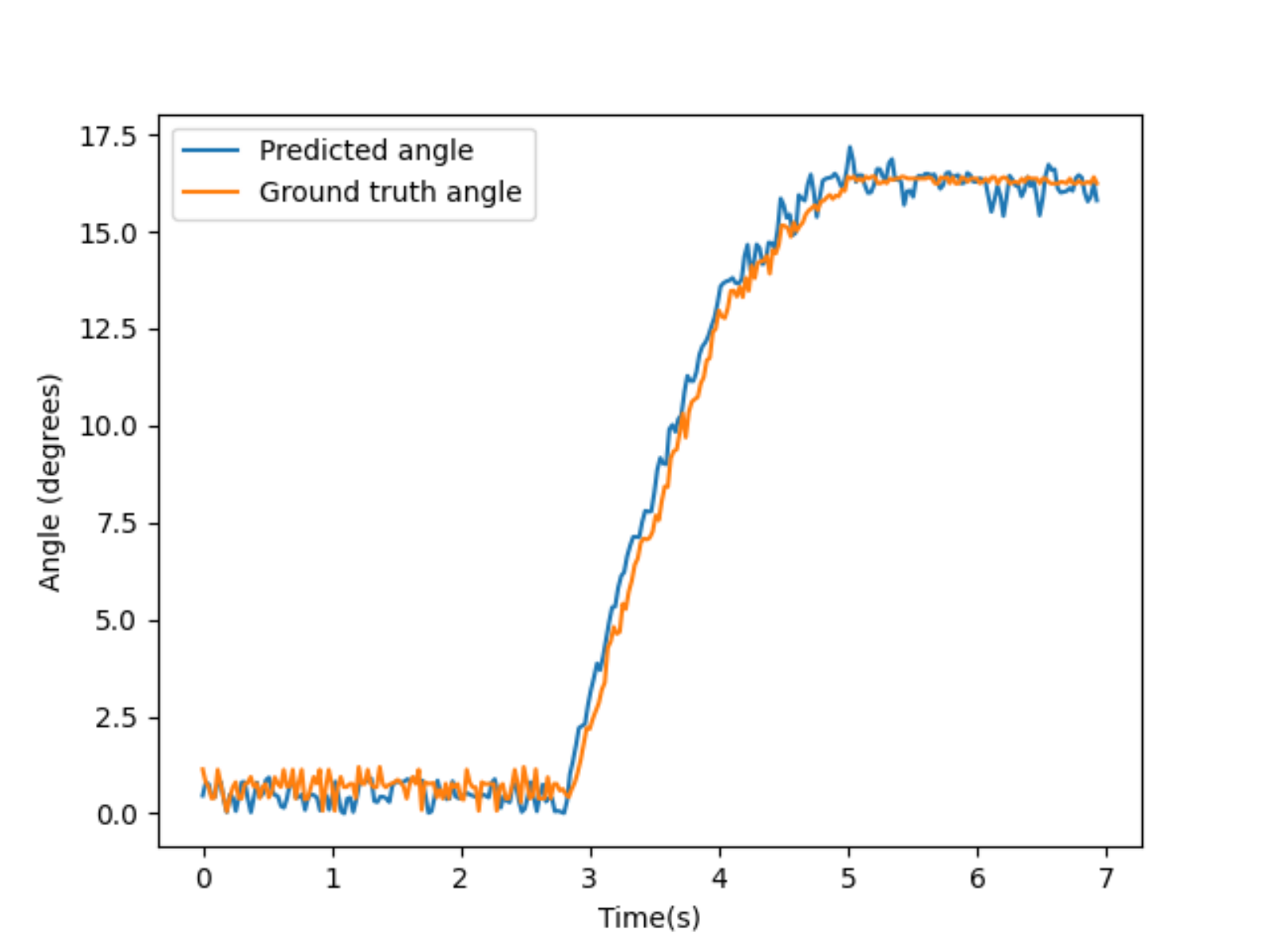}
         \caption{Object 5}
     \end{subfigure}

     \begin{subfigure}[b]{0.22\textwidth}
         \centering
         \includegraphics[height=3cm]{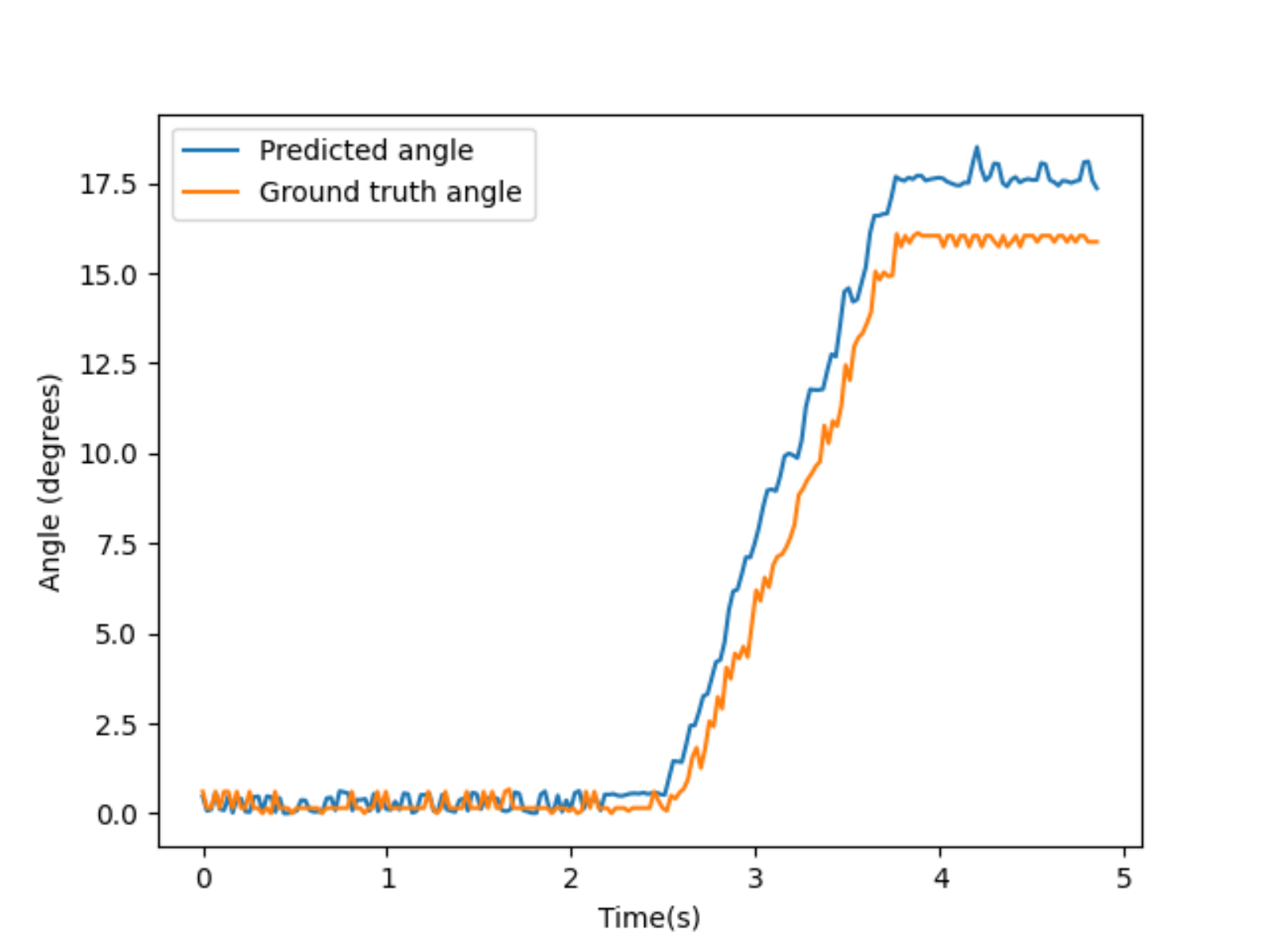}
         \caption{Object 4}
     \end{subfigure}
     \begin{subfigure}[b]{0.22\textwidth}
         \centering
         \includegraphics[height=3cm]{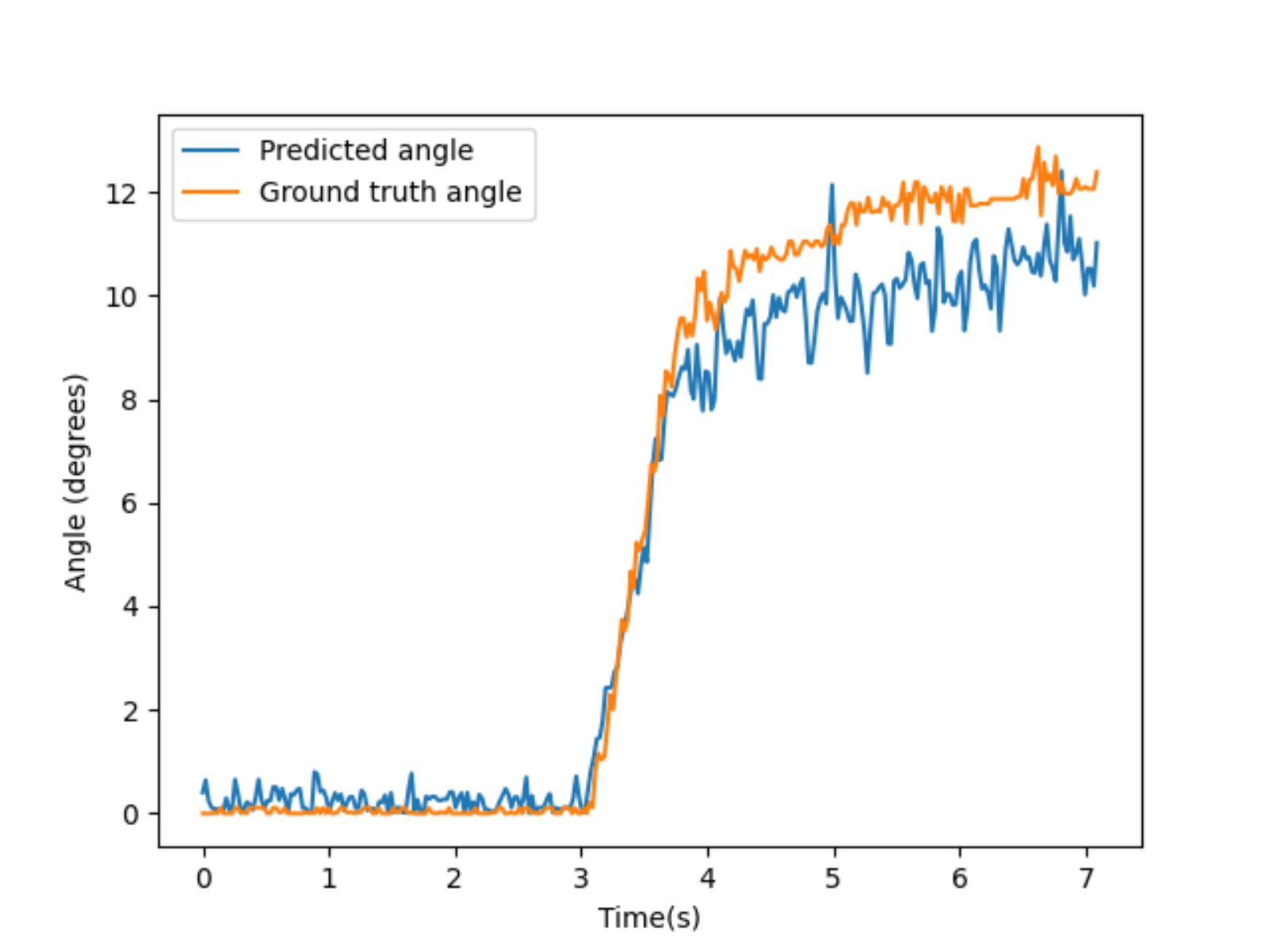}
         \caption{Object 1}
     \end{subfigure}

     \caption{Examples of rotation angle calculation for slipping during lift task with different objects}
      
     \label{fig:examples_angle_calculation}
\end{figure}

The experimentation comprises 45 graspings and lifts in total (five per object) while calculating the rotational error in degrees. Figure \ref{fig:skeleton_best_method} shows the mean rotational error of the 5 graspings and lifts for each object. Note that object 6 and 8 causes more error and deviation because object 6 weight's is higher compared with the rest of the objects, and object 8 contains higher curvature on its surface that causes more saturation in the sensor. Our system is able to predict rotational slippage with an overall mean rotational error of \textbf{1.854º $\pm$ 0.988º}, that is to say, a maximum mean error of 3 degrees in the worst case.  Figure \ref{fig:examples_angle_calculation} shows some examples of the prediction of rotational slippage with four aforementioned objects.


\section{Conclusions}
In this paper, we propose a model-based system to predict rotational slippage during the grasping and lift of an object, achieving a mean error value of \textbf{1.854º $\pm$ 0.988º}, compared with the error of \textbf{3.96º $\pm$ UNK} from \cite{kolamuri}, and the error of \textbf{4.39º $\pm$ 0.18º} from \cite{toskov}. Although we could not carry out an experimental comparison because we do not have their sensors available, some objects were used both in this work and in theirs. 
Our system also has some limitations regarding the shape of the contact region. If this shape is similar to a circle, it becomes impossible to calculate its rotation movement. In that case, we propose to grasp the object by surfaces with small curvature. 

\bibliography{sn-bibliography.bib}{}

\begin{thebibliography}{10}
\providecommand{\url}[1]{#1}
\csname url@rmstyle\endcsname
\providecommand{\newblock}{\relax}
\providecommand{\bibinfo}[2]{#2}
\providecommand\BIBentrySTDinterwordspacing{\spaceskip=0pt\relax}
\providecommand\BIBentryALTinterwordstretchfactor{4}
\providecommand\BIBentryALTinterwordspacing{\spaceskip=\fontdimen2\font plus
\BIBentryALTinterwordstretchfactor\fontdimen3\font minus
  \fontdimen4\font\relax}
\providecommand\BIBforeignlanguage[2]{{%
\expandafter\ifx\csname l@#1\endcsname\relax
\typeout{** WARNING: IEEEtran.bst: No hyphenation pattern has been}%
\typeout{** loaded for the language `#1'. Using the pattern for}%
\typeout{** the default language instead.}%
\else
\language=\csname l@#1\endcsname
\fi
#2}}

\bibitem{du}
G.~Du, K.~Wang, S.~Lian, and K.~Zhao, ``Vision-based robotic grasping from
  object localization, object pose estimation to grasp estimation for parallel
  grippers: a review,'' \emph{Artificial Intelligence Review}, vol.~54, no.~3,
  pp. 1677--1734, 2021.

\bibitem{luo}
S.~Luo, J.~Bimbo, R.~Dahiya, and H.~Liu, ``Robotic tactile perception of object
  properties: A review,'' \emph{Mechatronics}, vol.~48, pp. 54--67, 2017.

\bibitem{chi}
C.~Chi, X.~Sun, N.~Xue, T.~Li, and C.~Liu, ``Recent progress in technologies
  for tactile sensors,'' \emph{Sensors}, vol.~18, no.~4, p. 948, 2018.

\bibitem{hardwaretactil}
S.~Zhang, Z.~Chen, Y.~Gao, W.~Wan, J.~Shan, H.~Xue, F.~Sun, Y.~Yang, and
  B.~Fang, ``Hardware technology of vision-based tactile sensor: A review,''
  \emph{IEEE Sensors Journal}, 2022.

\bibitem{digit}
M.~Lambeta, P.-W. Chou, S.~Tian, B.~Yang, B.~Maloon, V.~R. Most, D.~Stroud,
  R.~Santos, A.~Byagowi, G.~Kammerer, \emph{et~al.}, ``Digit: A novel design
  for a low-cost compact high-resolution tactile sensor with application to
  in-hand manipulation,'' \emph{IEEE Robotics and Automation Letters}, vol.~5,
  no.~3, pp. 3838--3845, 2020.

\bibitem{pytouch}
M.~Lambeta, H.~Xu, J.~Xu, P.-W. Chou, S.~Wang, T.~Darrell, and R.~Calandra,
  ``Pytouch: A machine learning library for touch processing,'' in \emph{2021
  IEEE International Conference on Robotics and Automation (ICRA)}.\hskip 1em
  plus 0.5em minus 0.4em\relax IEEE, 2021, pp. 13\,208--13\,214.

\bibitem{ito}
Y.~Ito, Y.~Kim, and G.~Obinata, ``Contact region estimation based on a
  vision-based tactile sensor using a deformable touchpad,'' \emph{Sensors},
  vol.~14, no.~4, pp. 5805--5822, 2014.

\bibitem{gelsight}
S.~Wang, Y.~She, B.~Romero, and E.~Adelson, ``Gelsight wedge: Measuring
  high-resolution 3d contact geometry with a compact robot finger,'' in
  \emph{2021 IEEE International Conference on Robotics and Automation
  (ICRA)}.\hskip 1em plus 0.5em minus 0.4em\relax IEEE, 2021, pp. 6468--6475.

\bibitem{bauza}
M.~Bauza, O.~Canal, and A.~Rodriguez, ``Tactile mapping and localization from
  high-resolution tactile imprints,'' in \emph{2019 International Conference on
  Robotics and Automation (ICRA)}.\hskip 1em plus 0.5em minus 0.4em\relax IEEE,
  2019, pp. 3811--3817.

\bibitem{lepora}
Y.~Lin, J.~Lloyd, A.~Church, and N.~F. Lepora, ``Tactile gym 2.0: Sim-to-real
  deep reinforcement learning for comparing low-cost high-resolution robot
  touch,'' \emph{IEEE Robotics and Automation Letters}, vol.~7, no.~4, pp.
  10\,754--10\,761, 2022.

\bibitem{riai}
J.~Casta{\~n}o-Amor{\'o}s, I.~d.~L. P{\'a}ez-Ubieta, P.~Gil, and S.~T. Puente,
  ``Manipulaci{\'o}n visual-t{\'a}ctil para la recogida de residuos
  dom{\'e}sticos en exteriores,'' \emph{Revista Iberoamericana de
  Autom{\'a}tica e Inform{\'a}tica industrial}, vol.~20, no.~2, pp. 163--174,
  2023.

\bibitem{brayan}
B.~S. Zapata-Impata, P.~Gil, and F.~Torres, ``Learning spatio temporal tactile
  features with a convlstm for the direction of slip detection,''
  \emph{Sensors}, vol.~19, no.~3, p. 523, 2019.

\bibitem{kolamuri}
R.~Kolamuri, Z.~Si, Y.~Zhang, A.~Agarwal, and W.~Yuan, ``Improving grasp
  stability with rotation measurement from tactile sensing,'' in \emph{2021
  IEEE/RSJ International Conference on Intelligent Robots and Systems
  (IROS)}.\hskip 1em plus 0.5em minus 0.4em\relax IEEE, 2021, pp. 6809--6816.

\bibitem{toskov}
J.~Toskov, R.~Newbury, M.~Mukadam, D.~Kulic, and A.~Cosgun, ``In-hand
  gravitational pivoting using tactile sensing,'' in \emph{Conference on Robot
  Learning}.\hskip 1em plus 0.5em minus 0.4em\relax PMLR, 2023, pp. 2284--2293.

\bibitem{deeplabv3+}
L.-C. Chen, Y.~Zhu, G.~Papandreou, F.~Schroff, and H.~Adam, ``Encoder-decoder
  with atrous separable convolution for semantic image segmentation,'' in
  \emph{Proceedings of the European conference on computer vision (ECCV)},
  2018, pp. 801--818.

\bibitem{skeleton}
Z.~Guo and R.~W. Hall, ``Parallel thinning with two-subiteration algorithms,''
  \emph{Communications of the ACM}, vol.~32, no.~3, pp. 359--373, 1989.

\bibitem{unet++}
Z.~Zhou, M.~M. Rahman~Siddiquee, N.~Tajbakhsh, and J.~Liang, ``Unet++: A nested
  u-net architecture for medical image segmentation,'' in \emph{Deep Learning
  in Medical Image Analysis and Multimodal Learning for Clinical Decision
  Support: 4th International Workshop, DLMIA 2018, and 8th International
  Workshop, ML-CDS 2018, Held in Conjunction with MICCAI 2018, Granada, Spain,
  September 20, 2018, Proceedings 4}.\hskip 1em plus 0.5em minus 0.4em\relax
  Springer, 2018, pp. 3--11.

\bibitem{pspnet}
H.~Zhao, J.~Shi, X.~Qi, X.~Wang, and J.~Jia, ``Pyramid scene parsing network,''
  in \emph{Proceedings of the IEEE conference on computer vision and pattern
  recognition}, 2017, pp. 2881--2890.

\end{thebibliography}
\bibliographystyle{IEEEtran.bst}

\end{document}